\pdfoutput=1

\documentclass[11pt]{article}

\usepackage[final]{acl}

\usepackage{times}
\usepackage{latexsym}

\usepackage[T1]{fontenc}

\usepackage[utf8]{inputenc}

\usepackage{microtype}

\usepackage{inconsolata}

\usepackage{graphicx}
\usepackage{listings}
\usepackage{hyperref}
\usepackage{multirow}
\usepackage{pdfpages}
\usepackage{url}
\usepackage{xcolor}
\usepackage{epsfig}
\usepackage{adjustbox}
\usepackage{amsfonts}
\usepackage{amsmath}
\usepackage{amssymb}
\usepackage{booktabs} 
\usepackage{comment}
\usepackage{caption}
\usepackage{subcaption}
\usepackage{textcomp}
\usepackage{relsize}
\usepackage{stmaryrd}
\usepackage{bbm}
\usepackage{rotating}
\usepackage{tcolorbox}
\usepackage{helvet}
\usepackage{courier}
\usepackage{natbib}
\usepackage{cleveref}
\usepackage{xspace}
\usepackage{enumitem}
\usepackage{makecell}
 \usepackage{array}
 \usepackage{threeparttable}
 \usepackage{xcolor}
 \usepackage[normalem]{ulem}

\newcommand{\eg}{{e.g.},~}%
\newcommand{\ie}{{i.e.},~}%
\lstdefinestyle{python}{
    language=Python,
    basicstyle=\fontsize{8}{10}\ttfamily,
    keywordstyle=\color{blue},
    commentstyle=\color{gray},
    stringstyle=\color{black},
    showstringspaces=false,
    breaklines=true,
    breakindent=0pt,
    breakatwhitespace=false,
    escapeinside={(*@}{@*)}
}

\lstdefinestyle{cpp}{
    language=C++,
    basicstyle=\fontsize{8}{10}\ttfamily,
    keywordstyle=\color{blue},
    commentstyle=\color{gray},
    stringstyle=\color{green},
    showstringspaces=false,
    breaklines=true,
    breakindent=0pt,
    breakatwhitespace=false,
    escapeinside={(*@}{@*)}
}

\lstdefinestyle{plain}{
    basicstyle=\fontsize{8}{10}\ttfamily,
    keywordstyle=\color{blue},
    commentstyle=\color{gray},
    stringstyle=\color{green},
    showstringspaces=false,
    breaklines=true,
    breakatwhitespace=false,
    breakindent=0pt,
    escapeinside={(*@}{@*)}
}

\lstdefinestyle{python2}{
    language=Python,
    basicstyle=\fontsize{8}{10}\ttfamily,
    keywordstyle=\color{blue},
    commentstyle=\color{gray},
    stringstyle=\color{green},
    showstringspaces=false,
    breakatwhitespace=false,
    breaklines=true,
    breakindent=0pt,
    escapeinside={(*@}{@*)}
}

\lstdefinestyle{cpp2}{
    language=C++,
    basicstyle=\fontsize{8}{10}\ttfamily,
    keywordstyle=\color{blue},
    commentstyle=\color{gray},
    stringstyle=\color{green},
    showstringspaces=false,
    breaklines=true,
    breakindent=0pt,
    breakatwhitespace=false,
    escapeinside={(*@}{@*)}
}

\lstdefinestyle{sql}{
    language=SQL,
    basicstyle=\fontsize{8}{10}\ttfamily,
    keywordstyle=\color{blue},
    commentstyle=\color{green},
    stringstyle=\color{black},
    showstringspaces=false,
    breakatwhitespace=false,
    breaklines=true,
    breakindent=0pt,
    escapeinside={(*@}{@*)}
}

\lstdefinestyle{prompt}{
    language=Python,
    basicstyle=\fontsize{8}{10}\ttfamily,
    keywordstyle=\color{blue},
    commentstyle=\color{gray},
    stringstyle=\color{cppgreen},
    showstringspaces=false,
    breaklines=true,
    backgroundcolor=\color{bgcolor},
    keepspaces=true, 
    breakindent=0pt,
    breakatwhitespace=false,
    showspaces=false,   
    escapeinside={(*@}{@*)}
}
\lstdefinestyle{text}{
    basicstyle=\fontsize{8}{10}\ttfamily,
    showstringspaces=false,
    breaklines=true,
    backgroundcolor=\color{bgcolor},
    breakatwhitespace=false,
    breakindent=0pt,
    keepspaces=true,
    showspaces=false,   
    escapeinside={(*@}{@*)}
}

\tcbset{
  aibox/.style={
    width=\textwidth,
    top=10pt,
    colback=white,
    colframe=black,
    colbacktitle=black,
    center title,
    fonttitle=\bfseries,
  }
}

\newtcolorbox{AIbox}[2][]{aibox,title=#2,#1}

\tcbset{
  aiboxsmall/.style={
    width=0.48\textwidth,
    top=10pt,
    colback=white,
    colframe=black,
    colbacktitle=black,
    center title,
    fonttitle=\bfseries,
  }
}

\newtcolorbox{AIboxSmall}[2][]{aiboxsmall,title=#2,#1}

\PassOptionsToPackage{table}{xcolor}

\title{One Missing Piece for Open-Source Reasoning Models: A Dataset to Mitigate Cold-Starting Short CoT LLMs in RL}

\author{%
    Hyungjoo Chae$^{1,2,*}$, 
    Dongjin Kang$^{1,2,*}$,
    Jihyuk Kim$^{2}$,
    Beong-woo Kwak$^{1}$, \\
    \textbf{Sunghyun Park}$^{2}$,
    \textbf{Haeju Park}$^{2}$,
    \textbf{Jinyoung Yeo}$^{1}$,
    \textbf{Moontae Lee}$^{2,3}$,
    \textbf{Kyungjae Lee}$^{2}$\\
    \\
    $^1$Yonsei University \quad
    $^2$LG AI Research \quad
    $^3$University of Illinois Chicago \\
    \texttt{\{mapoout, hard1010, jinyeo\}@yonsei.ac.kr}\\
    \texttt{\{moontae.lee, kyungjae.lee\}@lgresearch.ai}\\
\vspace{0.0em}
}
\begin{document}

\maketitle

\renewcommand{\thefootnote}{\fnsymbol{footnote}}
\footnotetext[1]{Equal contribution. Work was done during internship at LG AI Research.}

\renewcommand{\thefootnote}{\arabic{footnote}}

\begin{abstract}
With the release of R1, a publicly available large reasoning model (LRM), researchers commonly train new LRMs by training language models on R1's long chain-of-thought (CoT) inferences. While prior works show that LRMs' capabilities can be reproduced through direct distillation, the continued reliance on the existing models (\eg R1) remains a critical limitation in advancing the field.
As a first step toward independent LRM development, this paper explores the possibility of constructing a long CoT dataset with LLMs that are not trained for inference-time scaling.
To this end, we present the Long CoT Collection, a dataset of 100K CoT rationales annotated using existing short CoT LLMs. 
We develop a pipeline that induces o1's novel reasoning strategies into short CoT LLMs, enabling them to think longer and introducing controllability over the thought budget to better manage the overthinking problem.
Our extensive analyses validate that our dataset achieves quality comparable to\textemdash or slightly below\textemdash R1. 
Furthermore, our experiments demonstrate that training on our dataset not only strengthens general reasoning skills, but also provides a strong foundation for reinforcement learning\textemdash models initialized on our data achieve 2-3x larger gains with RLVR. 
We make the codes, datasets, and models publicly available at \href{https://bit.ly/long-cot-collection-github}{LINK}.
\end{abstract}
\section{Introduction}

\begin{figure}[t]
    \centering
    \includegraphics[width=1.0\linewidth]{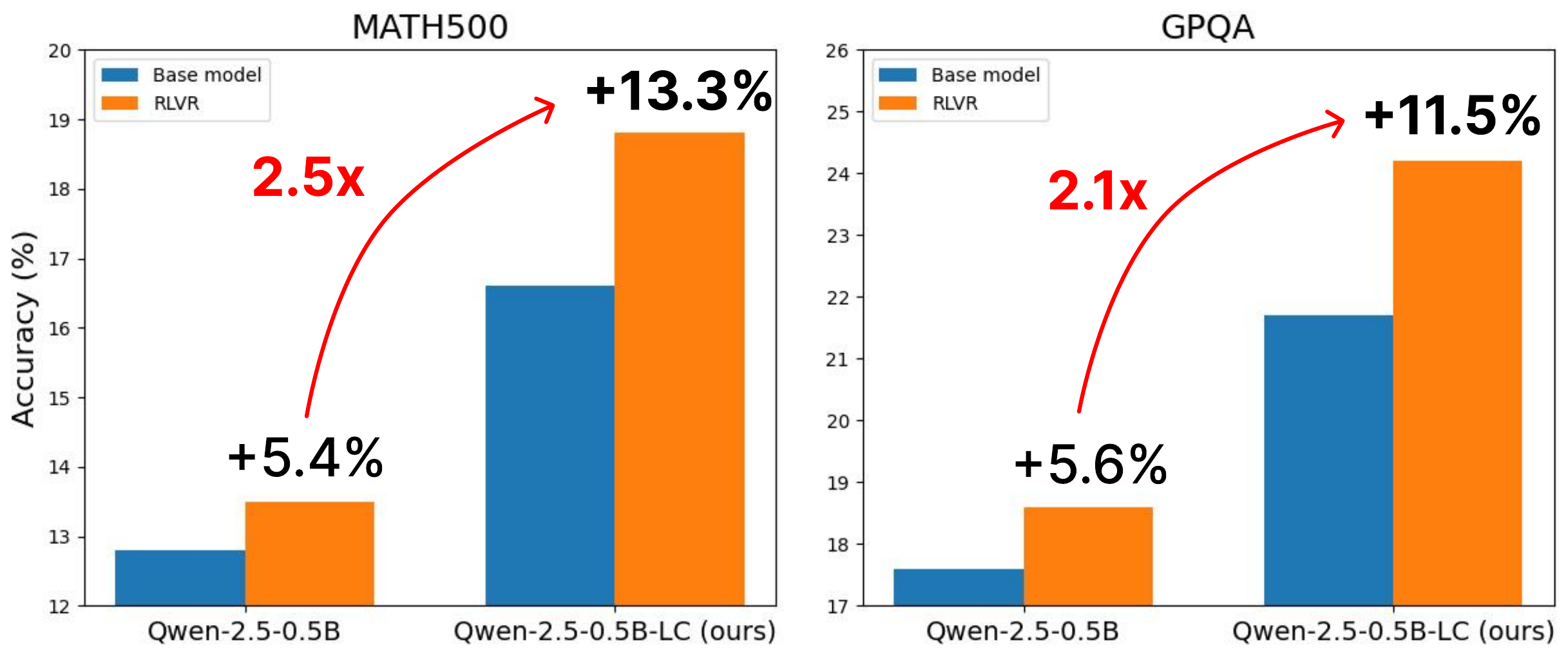}
    \caption{Comparison of RLVR performance between the base model (Qwen-2.5-0.5B) and the model trained on the Long CoT Collection (Qwen-2.5-0.5B-LC) on MATH500 and GPQA. }
    \label{fig:rlvr_results}
\end{figure}

Large Reasoning Models (LRMs), exemplified by the o-series~\citep{o1}, have shown groundbreaking performance in various reasoning tasks with test-time scaling (\ie generating extremely long chain-of-thought (CoT) rationales).
~\citep{rstarmath, o1coder, finemedlm-o1}.
However, their closed nature presents significant challenges\textemdash its high API costs and safety issues limit real-world applications~\citep{hendrycks2022unsolved}, while the closed-source approach potentially prohibits academic progress in the field. 

To address these issues, \citet{r1} release an open-source version of o1 and detail their methodology for building R1.
While the benefits of reinforcement learning with verifiable reward (RLVR) have been previously demonstrated~\citep{lambert2024t}, they introduce a key innovation by tackling the cold-start instability in RL training for Short CoT LLMs.
Finetuning on a carefully curated Long CoT dataset to explicitly teach reasoning structures serves as a critical step to enable the model to acquire 
the foundational reasoning skills before RL. Building on this insight, subsequent works have shown that simply collecting R1’s outputs to construct a Long CoT dataset and fine-tuning LLMs on it can lead to dramatic improvements~\citep{bespoke_stratos, OpenThoughts}. 
Furthermore, \citet{yeo2025demystifying} provide a detailed analysis of the role of RLVR following this finetuning stage.

Yet, despite these advancements, an important gap remains: the cold-start problem itself has not been fully demystified. While R1’s Long CoT dataset serves as a critical ingredient, the exact mechanisms for creating such data have remained unclear.
In this work, we investigate whether it is possible to construct Long CoT data from the short CoT responses of LLMs that have been trained to produce only concise rationales. 
Instead of directly collecting LRMs' completions, we built a simple pipeline that enables LLMs to generate long CoT in a step-by-step manner with only a small guidance from LRMs.
To allow LLMs to annotate long CoT, we begin by creating a seed dataset of 1K instances, capturing o1's reasoning flow that reflects its novel reasoning strategies. 
Then, we generate the reasoning flow on the new question and expand it to long CoT with short CoT LLMs (e.g., GPT-4o) in a step-by-step manner.
The resulting collection of 100K instances serves as a comprehensive training resource, allowing base LLMs to learn to think longer while incorporating diverse reasoning strategies characteristic of o1.
Since this collection process offers controllability over the thought budget, it has a strong advantage in addressing one of the major issues with LRMs: overthinking\textemdash generating an unnecessarily large number of tokens for simple problems.

To further validate our approach, we conduct in-depth analyses of the quality of our dataset.
Despite being generated by short CoT LLMs, the rationales in our dataset demonstrate reasoning flows and strategies that nearly match the quality of R1 in terms of reasoning flow, showing only slightly lower performance in other criteria. In addition, the generated rationals contain rich reasoning triggers (\eg ``Wait'' and ``To verify'') that help explore diverse reasoning paths and enhance accuracy. Our thought budget analysis shows that short CoT LLMs, guided by the example reasoning flow, effectively allocate their computational resources in alignment with state-of-the-art reasoning models.

Through extensive experiments, we demonstrate that the Long CoT Collection provides an effective foundation for initializing SFT models for reinforcement learning (RL).
Best-of-$n$ sampling comparisons show that models trained on our dataset consistently outperform the base models, demonstrating strong potential when optimized for outcome-based rewards.
Evaluations on GPQA~\citep{gpqa} and MMLU-Pro~\citep{wang2024mmlu} further highlight that training on our dataset enhances reasoning capabilities across general domain tasks.
Notably, initializing policies with our dataset before RL leads to 2-3x greater performance improvements, demonstrating out collection's strong potential to accelerate and stabilize downstream learning (Figure~\ref{fig:rlvr_results}).

%

\section{Related Work}
\paragraph{Inference-time Scaling.}

Recent research has demonstrated that scaling inference-time improves efficiency and overall reasoning quality by increasing the number of tokens, compared to traditional scaling laws such as increasing model parameters or dataset volumns~\citep{brown2024largelanguagemonkeysscaling, snell2024scalingllmtesttimecompute}. 
This can be achieved by sampling many reasoning paths (\eg Best-of-N~\citep{snell2024scalingllmtesttimecompute} and MCTS~\citep{zhang2024llama}) and using a verifier or voting mechanism to pick the correct solution (\eg self-consistency)~\citep{liang2024improving}.
Furthermore, \citet{o1, r1} explore training LLMs to generate a long CoT, similar to how humans handle complex tasks, which often involve self-correction or verification before arriving at a final answer.
This shift towards deliberative reasoning makes LLMs more transparent, interpretable, and adaptable in complex decision-making scenarios~\citep{yeo2025demystifying}.


\paragraph{Large Reasoning Models and Datasets.}
%
%

Since the success of OpenAI's o1 model~\citep{o1}, many studies have attempted to replicate o1-like reasoning as open-source models~\citep{qwq-32b-preview, sky_t1, muennighoff2025s1}.
Recent studies emphasize the importance of the dataset used for initializing these LRMs~\citep{xu2025redstardoesscalinglongcot, muennighoff2025s1, ye2025limo}.
Notably, \citet{r1} demonstrated that introducing a brief supervised fine-tuning (SFT) stage\textemdash where the model is ``cold-started'' with a few thousand high-quality CoT examples\textemdash leads to a more stable and efficient RL stage.
High-quality SFT datasets for reasoning are thus a key ingredient for these models, yet current public datasets remain limited.
To compensate, researchers have begun curating their own reasoning corpora~\citep{rstarmath, xu2025redstardoesscalinglongcot, pang2025bolt, ye2025limo}.
To address this critical gap, we introduce the Long CoT Collection, a large-scale dataset specifically designed to initialize models for complex reasoning tasks through supervised fine-tuning.

\begin{figure*}[!h]
    
    \centering
    \vspace{-6pt}
    \includegraphics[width=0.99\linewidth]{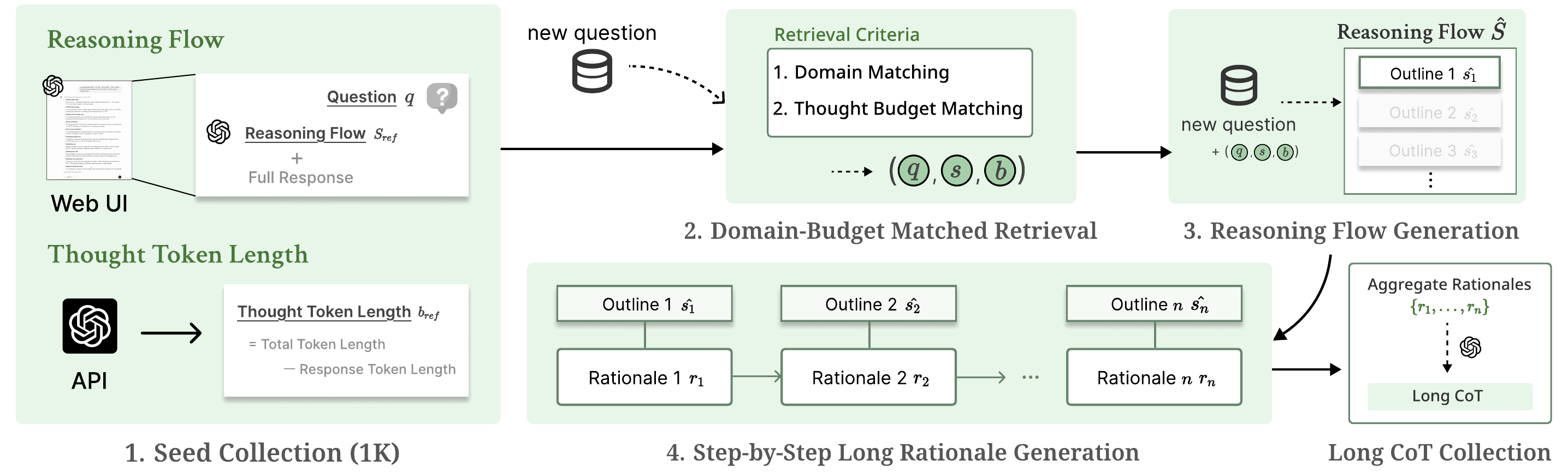}
    \caption{Overview of our data construction pipeline. First, we collect an 1K seed dataset of reasoning flow and thought token length (1). Using it as a demonstration, we annotate long CoT rationales on new questions and scale it up to 100K data points (2-4).}
    \vspace{-4pt}
    \label{fig:annotation_pipeline}
\end{figure*}

\paragraph{Reinforcement Learning for Reasoning.}
Reinforcement learning with human feedback (RLHF) has become a dominant paradigm for aligning LLMs to human preferences~\citep{ouyang2022traininglanguagemodelsfollow, bai2022traininghelpfulharmlessassistant, touvron2023llama}.
In RLHF, a reward model that learns human preference guides the policy to produce responses that humans would rate highly (\eg helpful and harmless responses)~\citep{zhu2023starling}.
However, \citet{lambert2024t} have pointed out that relying on a learned reward model can introduce instability in the RL process. 
To tackle this, researchers are turning to RLVR as a more grounded alternative for reasoning domains~\citep{lambert2024t, r1}.
The idea is to focus on objective, checkable outcomes rather than learning a proxy for human preferences, providing rewards only when its output is correct.

\section{The Long CoT Collection}

In this section, we present the Long CoT Collection, a dataset for learning LRMs' emergent reasoning behavior.
To allow more openness and controllability of the data collection process, we investigate whether long CoT data can be annotated by short CoT LLMs.
Our data collection process begins by collecting 1K demonstrations that capture LRMs' reasoning flow (Section~\ref{ssec:seed_data}), then generating 100K long CoT data using short CoT LLMs guided by the seed demonstrations (Section~\ref{ssec:expansion}).
The overall construction process is illustrated in Figure~\ref{fig:annotation_pipeline}.

\subsection{Collecting Teacher Demonstrations}
\label{ssec:seed_data}

A key challenge in building long CoT datasets with short CoT LLMs is allowing them to generate long rationales with coherence. To address this, we first collect a seed dataset with o1 that reflects the novel reasoning process of LRMs.

\subsubsection{Reasoning Flow Annotation}

\begin{figure}[!t]
    \centering
    \vspace{-6px}
    \includegraphics[width=1.0\linewidth]{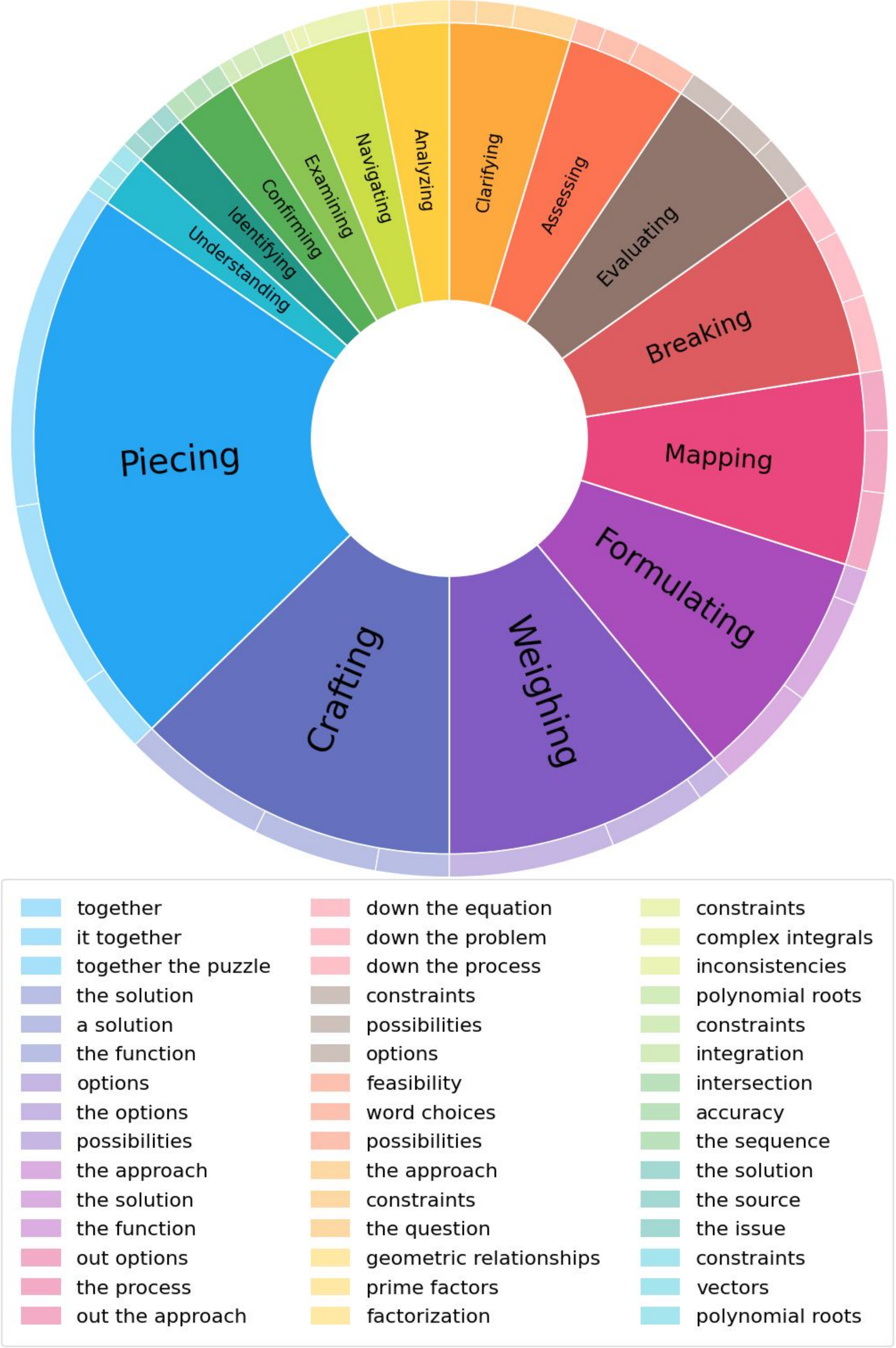}
    \caption{The top 15 most common root verbs and
their top 3 direct noun objects in the collected
reasoning flow.}
    \vspace{-4px}
    \label{fig:thought_summary_keyword}
\end{figure}

Reasoning flow $S$ is an overview of the reasoning process that consists of a sequence of outlines $\{s_1, s_2, ..., s_n\}$ for each reasoning step. It contains crucial information about the reasoning process and how the logical steps flow from the initial problem understanding to the final conclusion. We manually collect reference reasoning flow $S_{ref}$ from ChatGPT website, using the question $q$ from 1K reasoning-focused instructions from the magpie-reasoning-V1 dataset~\citep{xu2024magpie}. In addition, our dataset includes thought budget $b_{ref}$ (\ie the number of thought tokens used) of o1 by calculating the difference between the total completion token count and the number of tokens in the returned response, using the OpenAI API. As a result, we collect 1K seed dataset $\mathcal{D}_{ref} \in \{q, S_{ref}, b_{ref}\}$ that will be used in Section~\ref{ssec:expansion}. 
We show the distribution of the title of the reasoning outline in Figure~\ref{fig:thought_summary_keyword}.

\subsection{Annotating Long CoT with Indirect Guidance from Teacher}
\label{ssec:expansion}
Using the 1K seed dataset as our foundation, we expand it to 100K data. 
Since short CoT LLMs struggle to maintain coherence during extended test-time computing, we breakdown the reasoning into three steps to enable step-by-step generation of long CoT rationales.


\subsubsection{Reasoning Flow Retrieval}

Each question has its own reasoning procedure to reach the answer. Thus, for the new question $q$, we dynamically retrieve demonstrations $(q, S_{ref}, b_{ref})$ from our seed dataset $\mathcal{D}_{ref}$ to teach LLMs to generate reasoning flow $S$ with in-context learning. 
The following aspects are considered for the retrieval:
 \textbf{(1) Domain matching}: Problems in the same or similar domain are highly likely to share a common reasoning process. For example, in arithmetic reasoning, o1 tends to verify its calculation to ensure the correct answer. We use the primary domain and sub-domain in the magpie-V1-reasoning dataset to calculate the domain matching score~\citep{xu2024magpie}.
\textbf{(2) Thought budget control}: 
To align with reference LRMs, the thought budget is controlled by retrieving reasoning flows of similar length for demonstration. 
We measure this similarity using $1 - \left| \frac{\min(x, y)}{\max(x, y)} - 1 \right|$, where $x$ and $y$ represent the reference and candidate budgets, respectively. The heatmap of this similarity function is in Figure~\ref{fig:length_reward}.


\subsubsection{Reasoning Flow Generation}
\label{sssec:reasoning_flow_gen}
The retrieved demonstrations teach LLMs, which is GPT-4o in our experiment, to imagine LRMs' reasoning behavior at a higher level. Without the demonstration, we find that LLMs only stick to a linear thinking process, where the reasoning proceeds in one direction and does not include LRMs' novel reasoning strategies, such as verification and exploration of diverse solutions. LLMs generate reasoning flow $\hat{S}$ on the new question, given the retrieved demonstration. Specifically, they first predict the expected number of outlines $|S|$ and generate a sequence of reasoning outlines that emulates the higher-level reasoning patterns observed in the retrieved demonstrations.


\subsubsection{Step-by-step Long CoT Generation with Reasoning Flow}
\label{sssec:step_by_step}
Using the generated reasoning flow $\hat{S}$ as guidance, LLMs generate long CoT rationale step-by-step. Specifically, for each step $\hat{s}_i$ in $\hat{S}$ LLMs generate rationales $r_i$ based on the given previous reasoning $\{r_k\}_{0}^{i-1}$, the current flow step $\hat{s}_i$, and the next flow step $\hat{s}_{i+1}$. 
When the summary steps are all consumed, the LLMs generate the final solution based on the reasoning. 
At last, the reasoning steps and the final answer are aggregated as a sequence. 

\subsubsection{Correctness Filtering}
Lastly, we filter out the rationales that results in wrong answers, as training on incorrect rationales might harm their original reasoning capability. Specifically, we simply ask GPT-4o to validate the answer given the reference answer and the generated answer span. This filtering results in 76\% instances with correct answer prediction.



\section{Dataset Analyses}
\subsection{High Quality}
\label{ssec:high_quality}
\begin{figure}[t]
    \centering
    \vspace{-6px}
    \includegraphics[width=0.9\linewidth]{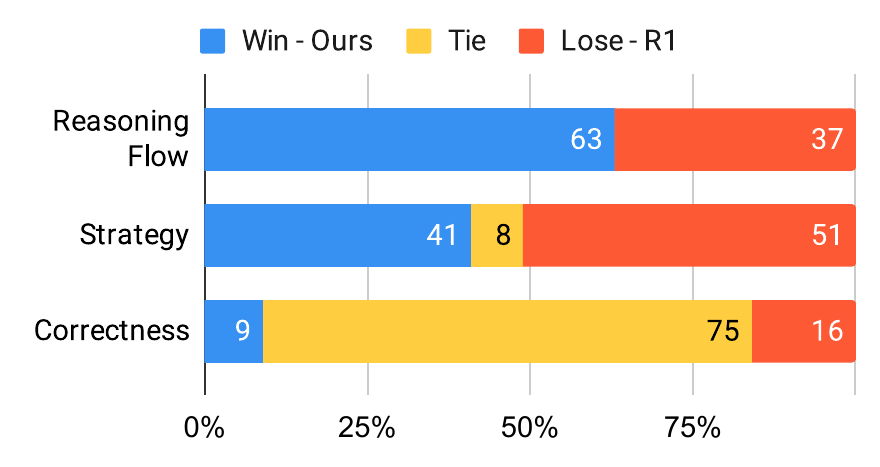}
    \caption{Head-to-head comparison of the generated CoT quality with the  R1 output~\citep{ye2025limo}.}
    \vspace{-4px}
    \label{fig:data_quality_magpie}
\end{figure}
We focus on three important aspects; (1) \textbf{Reasoning Flow}: The logical progression and coherence of steps in the solution process, measuring how naturally one step leads to the next. (2) \textbf{Reasoning Strategy}: The specific techniques and approaches employed to break down and solve problems, such as the selection of relevant mathematical tools or problem-solving methods. (3) \textbf{Correctness}: The accuracy of each reasoning steps. 

We compare our method with a widely used method for long CoT data generation which collects the outputs from the existing LRMs. 
For a fair comparison, we sample 100 questions from the Long CoT Collection for which R1-generated solutions have the correct answer.
Following the finding that stronger policy models can be used for trajectory scoring~\citep{wang2024q}, we use the state-of-the-art LRM, o3-mini, as our evaluator. Figure~\ref{fig:data_quality_magpie} shows that the rationales from the Long CoT Collection demonstrate better reasoning flow, and while showing slightly weaker strategy and correctness, they remain competitive. 

\subsection{Efficient Thought Budget Allocation}
\begin{figure}[t]
    \centering
    \includegraphics[width=.8\linewidth]{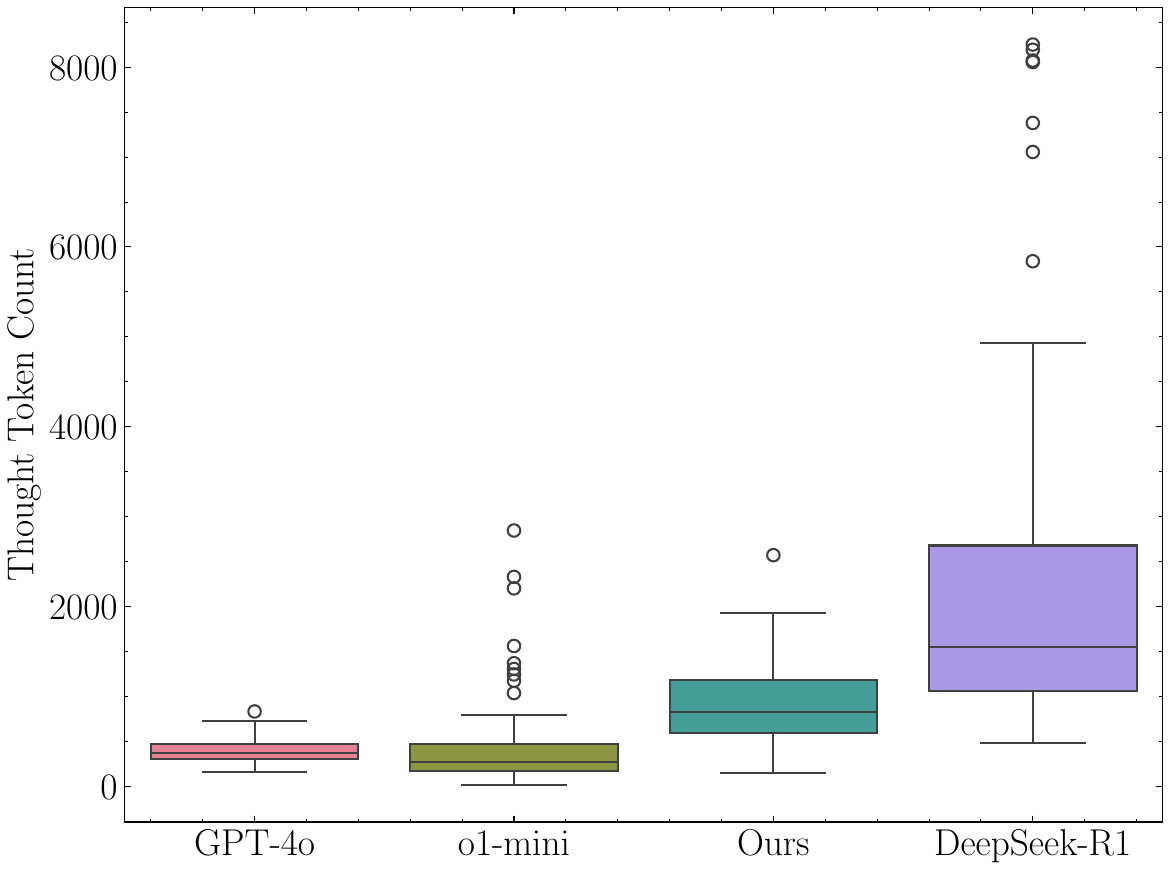}
    \caption{Comparison of the number of reasoning tokens used by each model.}
    \label{fig:thought_budget}
\end{figure}

Allocating the proper budget for thinking is an important issue~\citep{wang2025thoughts}. LRMs tend to use too many thought tokens for easy problems (\ie{} overthinking), which leads to a huge amount of computational cost.
To evaluate the efficiency in thought token allocation, we analyze the rationale lengths and compare them against other LRMs and GPT-4o, the LLM used in constructing our dataset.
Specifically, we randomly sample 100 instances from the Long CoT Collection and annotate the rationales with each model. As Figure~\ref{fig:thought_budget} indicates, simple CoT prompting on GPT-4o rarely generates rationales longer than 1,000 tokens, which suggests that naive prompting on GPT-4o is hard to use for constructing long CoT datasets. In addition, R1 uses significantly more thought tokens than o1-mini, which results in overthinking when models are trained on its outputs.

\begin{figure*}[!t]
    \centering
    \includegraphics[width=1.0\linewidth]{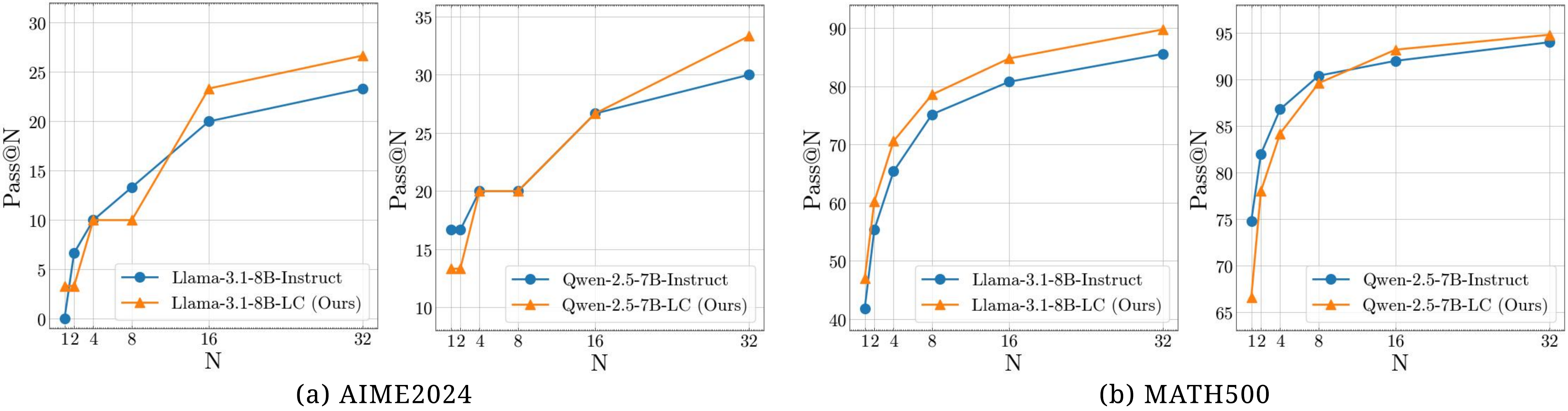}
    \vspace{-2em}
    \caption{Results of best-of-$n$ experiments with Llama-3.1-8B-Instruct and Qwen-2.5-7B-Instruct.}
    \label{fig:bon}
\end{figure*}

\section{Effect of the Long CoT Collection}
As demonstrated in prior works~\citep{r1,yeo2025demystifying}, the training of LRMs typically follows a two-phase approach: first, imitation learning to master long-form CoT reasoning, followed by RL to enhance reasoning accuracy.
In this section, we investigate the impact of training LLMs on our dataset from two perspectives: its effectiveness as a starting point for RL and its actual impact on the RL training phase.


\subsection{A Reliable Starting Point for RL}

\paragraph{Setup.}
RL for inference-time scaling includes sampling trajectories from the policy model and updating the policy based on calculated rewards. In such sparse reward settings, the quality of the initial policy model is critical\textemdash{}if the model rarely generates high-reward trajectories at the start, the learning signal may be too weak for effective training. To assess the potential of our initial policy model, we evaluate its performance using best-of-$n$ (BoN) sampling, which reveals the model's capacity to generate correct solutions when allowed multiple attempts. 
We assess our model on mathematical reasoning benchmarks,  as they widely used for RL to elicit inference-time scaling. We choose two challenging benchmarks, MATH-500~\citep{lightman2023letsverifystepstep} and AIME24~\citep{aime}.
We use Llama-3.1-8B-Instruct~\citep{dubey2024llama3herdmodels} and Qwen2.5-7B-Instruct~\citep{qwen2024qwen25technicalreport} as our base model. We train them on our dataset, and the full hyperparameters are in Appendix~\ref{app:sft}.

\paragraph{Results.}
Figure~\ref{fig:bon} shows BoN results with two base models. We measure Pass@$N$ ($N$=1,2,4,8,16 and 32), where a set of $N$ samples is considered correct if at least one sample includes the ground-truth answer.
On Llama-3.1-8B-Instruct, we observe notable improvement on both benchmarks, consistently across different $N$. 
Meanwhile, our Qwen2.5-7B-LC improves performance given large $N$ (e.g., 16 or 32), while the performance of Qwen2.5-7B-Instruct quickly saturates.
This shows that our SFT training recipe enables the model to explore more diverse responses and thus leads to higher answer reward when applied to RL.

\subsection{Impact on General Reasoning Domains}

\paragraph{Setup.}
Along with the mathematical benchmarks, we test our model on the general reasoning benchmarks, GPQA Diamond~\citep{gpqa} and MMLU-Pro~\citep{wang2024mmlu} (see Appendix~\ref{app:benchmark} for details).
We consider the baselines in the following three categories;
{(1) Closed-source LRMs}: OpenAI's o1 and o1-mini~\citep{o1} demonstrate state-of-the-art performance but are accessible only through APIs.
{(2) Open-source LRMs with undisclosed SFT datasets}: R1~\citep{r1} and QwQ~\citep{qwq-32b-preview} successfully replicate o1's capabilities, but the datasets for SFT remain undisclosed.
{(3) Open-source LRMs via distillation}: Models like Sky-T1~\citep{sky_t1} and Bespoke-7B~\citep{bespoke_stratos} utilize open-source datasets collected from existing LRMs' outputs.

\begin{table}[!tb]
\centering
\vspace{-6pt}
\resizebox{0.88\linewidth}{!}{%
\begin{tabular}{lccc}
\toprule
\textbf{Model} & \makecell{\textbf{Size}} & \makecell{\textbf{GPQA}\\\textbf{Diamond}} & \makecell{\textbf{MMLU}\\\textbf{Pro}} \\
\midrule
\multicolumn{4}{c}{\textbf{API only}} \\[-0.4ex]
\midrule
o1-mini & N/A & 60.0 & 80.3 \\
o1 & N/A & 77.3 & - \\
\midrule
\multicolumn{4}{c}{\textbf{Open Weights}} \\[-0.4ex]
\midrule
Qwen-2.5-32B-Instruct & 32B & 49.0 & 69.2 \\
QwQ-32B & 32B & 65.2 & 71.0 \\
R1 & 671B & 71.5 & 84.0 \\
Qwen-7B-R1-distill & 7B & 49.1 & - \\
\midrule
\multicolumn{4}{c}{\textbf{Open Weights and Open Data}} \\[-0.4ex]
\midrule
Sky-T1 & 32B & 56.8 & 69.2 \\
Bespoke-7B & 7B & 38.9 & - \\
OpenThinker-7B & 7B & 42.4 & - \\

\midrule
Llama-3.1-8B-Instruct & 8B & 22.7 & 43.7 \\
Llama-3.1-8B-LC (Ours) & 8B & 36.4 & 44.5 \\
Qwen-2.5-7B-Instruct & 7B & 37.6 & 49.9 \\
Qwen-2.5-7B-LC (Ours) & 7B & 39.9 & 51.4 \\
\bottomrule
\end{tabular}%
}
\caption{Performance of various reasoning models. Some results are from the respective reports
}
\vspace{-4pt}
\label{tab:main_table}
\end{table}
\paragraph{Results.}
We present our results in Table~\ref{tab:main_table}. Models trained on the Long CoT Collection show significant performance gains on GPQA, particularly Llama-3.1-8B-Instruct. Notably, Qwen-2.5-7B-LC achieves GPQA performance slightly surpassing Bespoke-7B, a simpler replication of R1. The models also demonstrate modest improvements on MMLU-Pro, suggesting that the reasoning strategies learned from our dataset transfer effectively to general reasoning domains.


\begin{figure*}[t]
    \centering
    \vspace{-6pt}
    \includegraphics[width=1.0\linewidth]{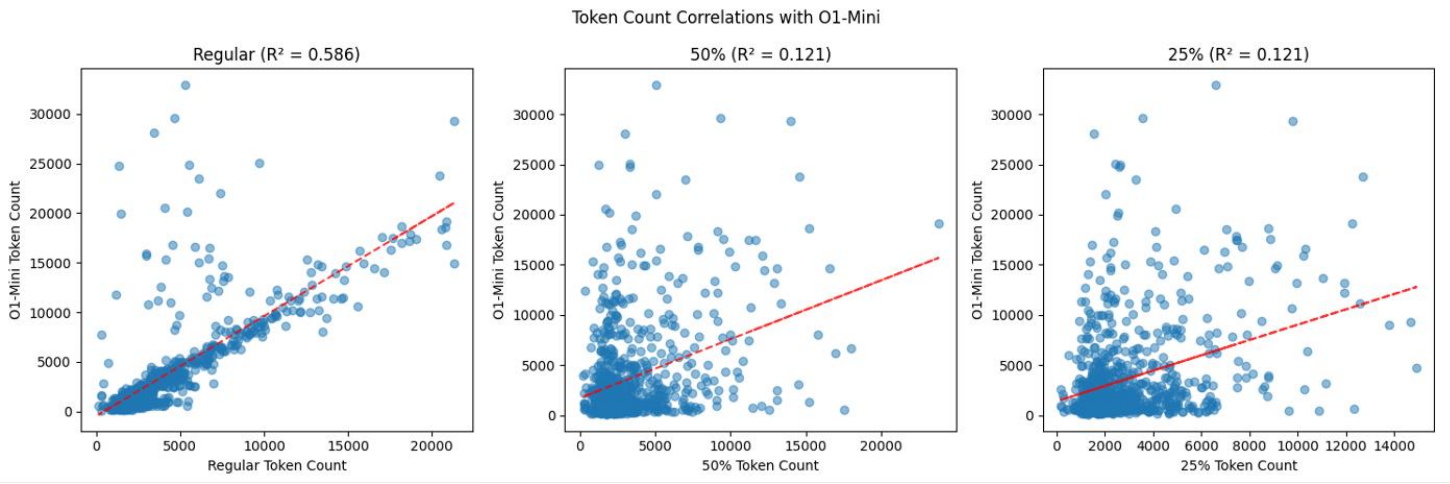}
    \caption{The Pearson correlation ($R^2$) between generated tokens and o1-mini thought tokens. We leverage 100\% budget (\textit{left}), 50\% budget (\textit{mid}), and 25\% budget (\textit{right}) to generate the collections of long CoT rationales.}
    \vspace{-4pt}
    \label{fig:corr_budget_controll}
\end{figure*}
\begin{figure}[t]
    \centering
    \includegraphics[width=1.0\linewidth]{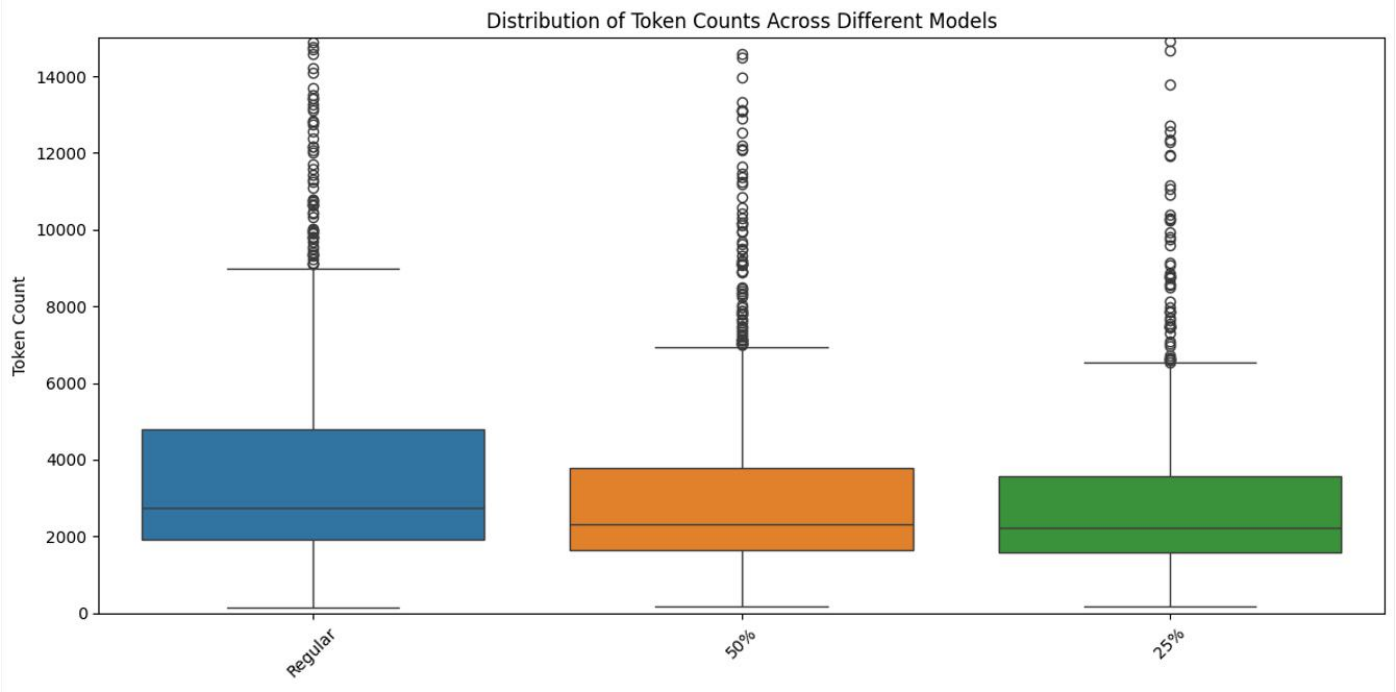}
    \vspace{-2em}
    \caption{The token count distribution for each collection generated by adjusting the thought budget.}
    \label{fig:dist_budget_controll}
\end{figure}

\subsection{Implication on RL}
\label{sec:implication_rl}

After imitation learning to develop the long-form CoT reasoning skills, we move on the next phase\textemdash RLVR with GRPO~\citep{deepseekmath}\textemdash to validate whether our collection serves as a reliable starting point for reinforcement learning.
Due to GPU resource constraints for long-sequence RL, we train Qwen-2.5-0.5B on the Long CoT Collection and leverage it as the starting point for RL.
Based on the NuminaMATH~\citep{numina_math_datasets}, we filter samples to include only those with integer answers, resulting in a set of 10K examples.
The policies are trained with a 16K max token length, using 16 samples per example for GRPO.
For verifiable rewards, following three types of reward functions are employed.

\paragraph{Reward Functions.}
There are three reward functions we employed, which are generally used for RL: 
(1) Length Reward: We use the function $1 - \left| \frac{\min(x, y)}{\max(x, y)} - 1 \right|$ that measures the difference between the length of sampled thought and o1-mini's thought on a scale of 0 to 1.
(2) Answer Reward: An outcome-based reward following \citet{yeo2025demystifying}. Specifically, we parse the answer span and compare it with the answer using latex2sympy,
(3) Format Reward: We check whether the model responses include the parable answer span.

\paragraph{Results.}
Figure~\ref{fig:rlvr_results} represents the impact of our Long CoT Collection on the next RL phase.
On both MATH500 and GPQA, the model initialized by training on our collection (\ie Qwen-2.5-0.5B-LC) achieves 2-3x greater performance gains through RLVR compared to the base model (\ie Qwen-2.5-0.5B), effectively mitigating the cold start problem.
This indicates that the Long CoT Collection serves as a reliable starting point for RL, showing the potential to enable more stable learning even under sparse reward signals and finally leading to greater performance gains.

\section{Thought Budget Control}
One of the major issues with long-sequence reasoning models is overthinking\textemdash generating an unnecessarily large number of tokens for simple problems.
For instance, QwQ-32B produces around 1,500 tokens for a basic question like '1+1+3?'.
Similarly, OpenAI's O-series models offer three types\textemdash low, medium, high\textemdash based on computational budget, allowing users to adjust the thinking budget according to the task complexity.
Now, having control on the thought budget is crucial for effectively managing the problem of overthinking.

\subsection{Controlling the Thought Budget During the Data Collection Process}
As described in Section~\ref{sssec:reasoning_flow_gen}, when synthesizing long rationles from short CoT LLMs, our collection process first generates reasoning outlines by estimating the number of outlines needed for each instance and then producing a sequence of reasoning outlines accordingly.
This allows us to control the length of the generated rationales by enforcing the number of outlines needed.
We finally craft three versions of the Long CoT Collection by additionally constructing two more sets, each constrained to use only 25\% and 50\% of the original budget.

\subsection{Analysis on the Budget-Controlled Collection}
Figure~\ref{fig:corr_budget_controll} illustrates the correlation ($R^2$) between the tokens of the generated rationales and o1-mini thought tokens.
It demonstrates that as we reduce the thought budget and generate relatively shorter rationales, the correlation with o1-mini thought tokens weakens.
Moreover, we figure out that excessively reducing the thought budget\textemdash specifically to 25\%\textemdash disrupts rationale generation by forcing too much information into too few reasoning outlines, making the reasoning more confusing.

We also investigate the distribution of each collection (\ie 100\%, 50\%, and 25\%).
As presented in Figure~\ref{fig:dist_budget_controll}, a reduction in the thought budget results in a corresponding decrease in the average token length of the collection.
Furthermore, policies trained with access to larger budgets exhibit superior reasoning capability compared to those trained under more constrained budgets (Table~\ref{tab:budget_controlled}).

\begin{table}[t]
\centering
\small
\resizebox{0.95\linewidth}{!}{
\begin{tabular}{l|ccc}
\toprule
\textbf{~~~~Data Used~~~~} & \textbf{100\%} & \textbf{50\%} & \textbf{25\%} \\
\midrule
~~~~MATH500~~~~ & ~~~~66.6~~~~ & ~~~~60.7~~~~ & ~~~~57.6~~~~ \\
\bottomrule
\end{tabular}
}
\caption{The results of policies on MATH500, which trained on each Long CoT Collection.}
\vspace{-10px}
\label{tab:budget_controlled}
\end{table}

\section{Conclusion}
This paper investigates the feasibility of generating long CoT datasets using LLMs trained on short CoT rationales. 
We present a pipeline for building the Long CoT Collection using short CoT LLMs, where the collection process offers controllability over the thought budget.
This gives us the ability to regulate the length of the generated rationales and provides a way to address overthinking\textemdash one of the major challenges faced by LRMs.
While training on our dataset did not lead to dramatic improvements over direct distillation from LRMs, our extensive experiments show that once moving into the RL phase, policies initialized with our dataset achieved 2-3x greater performance gains compared to those without it.
This highlights the strength of our dataset as an reliable foundation for RL.
\section*{Limitations and Future Work}
\paragraph{Application on Expert Domains.}
An exciting next step is to apply our pipeline to expert domains. 
While our dataset has proven to be a reliable starting point for RL in math and general reasoning tasks, we anticipate its potential to generalize further across a wide range of specialized domains.


\paragraph{Scaling Up to Larger Models.}
Although we employ 7B-8B models during phase 1 learning (\ie supervised fine-tuning), we use a 0.5B model for phase 2 (\ie reinforcement learning) since the largest model that fits within our GPU resources (16 A100 40GB GPUs) is 0.5B parameters. 


\paragraph{Using Diverse Teacher LRMs.}
We only consider o1 for the reference LRM used in our dataset construction process. While we choose o1 due to its representativeness, our approach can be further applied to other LRMs that partially disclose their reasoning processes.

\section*{Acknowledgements}
This work was supported by LG AI Research. Kyungjae Lee is the corresponding author.

\bibliography{custom}

\appendix

\section{Details of the Data Construction}
\label{app:data_construction}

\subsection{Base Dataset}
\citet{xu2024magpie} propose a set of synthetic instruction data, Magpie, which covers a wide range of domains.
From Magpie, a dataset consisting of 150K of the longest examples from reasoning, math, and coding \& debugging categories was also released.
Our 1K seed datasets and 100K long CoT collection are stems from the Magpie-Reasoning-150K.\footnote{\url{https://huggingface.co/datasets/Magpie-Align/Magpie-Reasoning-V1-150K}}
Each data point is annotated with multiple subcategories along with its main category.

\subsection{Details of Demonstration Retrieval}
\label{app:details_of_retrieval}
We leverage the main category and subcategories annotated in the dataset to retrieve demonstrations. We calculate the domain matching score by assigning 1 point for matching main categories and 0.2 points for each matching subcategory. The final retrieval score is computed by multiplying the domain matching score with the thought budget score, prioritizing samples with similar thought budgets.


\subsection{Statistics}
\begin{figure}[!t]
    \centering
    \includegraphics[width=1.0\linewidth]{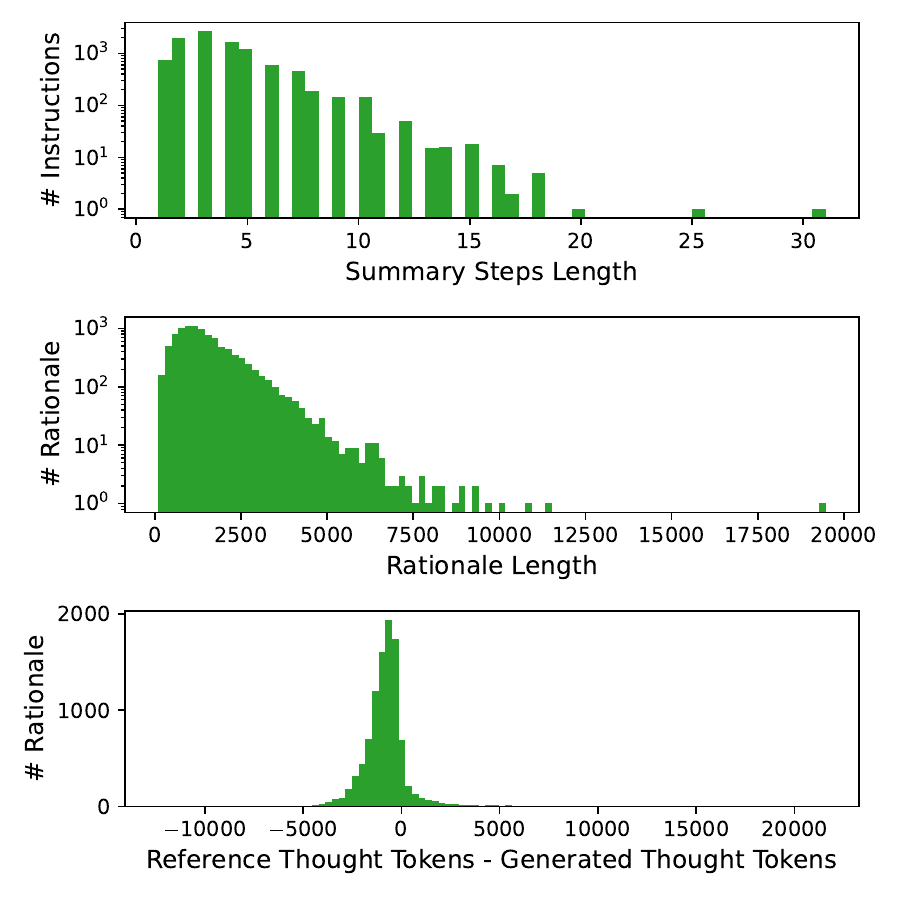}
    \caption{Distribution of the number of reasoning outlines (top), thought length (middle), and the difference between the length of the reference models' rationale and the generated rationale (bottom).}
    \label{fig:distribution}
\end{figure}

Figure~\ref{fig:distribution} shows the distributions of reasoning steps, rationale lengths, and differences between reference and generated thought tokens. Our dataset contains sufficiently long rationales, with up to 30 reasoning steps and 20K thought tokens. 
The comparison between generated and reference thought tokens reveals similar distributions, suggesting our approach may help prevent under- and over-thinking issues~\citep{wang2025thoughts} and enable short CoT LLMs to produce long rationales.

\begin{table}[!t]
\centering
\resizebox{0.8\linewidth}{!}{
\begin{tabular}{l|ccc}
\toprule
\textbf{Phrase}& \textbf{GPT-4o} & \textbf{Deepseek-R1} & \textbf{Ours} \\
\midrule
``Let's'' & 37 & 92 & 100 \\
``Wait'' & 0 & 100 & 4 \\
``Okay'' & 0 & 100 & 47 \\
``Verif-'' & 4 & 60 & 27 \\
``?'' & 0 & 87 & 27 \\
``!'' & 0 & 4 & 2 \\
\bottomrule
\end{tabular}
}
\caption{Frequency of ``Aha'' moment phrases in CoT rationales across different methods, representing the proportion (\%) of samples in which each phrase appears.}
\label{tab:aha}
\end{table}
\subsection{Analysis of Reasoning Triggers}
Prior work reports that LRMs frequently use ``Aha'' moment phrases to explore better reasoning paths~\citep{huang2024o1replicationjourney,r1,s1}.
These phrases serve not only as formatting elements but also as critical keywords that can steer the model's reasoning process, effectively guiding it towards more structured and thorough problem-solving. 
Thus, we check the frequency of these reasoning triggers, such as ``Let's think,'' ``Wait, I need to verify,'' and question marks indicating self-reflection. 
As shown in Table~\ref{tab:aha}, while GPT-4o exhibits minimal use of these markers (primarily ``Let's'' at 37\%), Deepseek-R1 employs them extensively across all categories. 

\section{Implementation Details}
\label{app:implementation_details}

\subsection{Datasets}
In Section~\ref{ssec:impact_vr}, LIMO~\citep{ye2025limo} dataset serves as a test-bed to assess the potential of our initialized SFT model in RL.
The LIMO dataset stems from NuminaMath-CoT, featuring meticulously annotated problems from high school to advanced competition levels, AIME, and MATH.
It contains 817 meticulously selected math problems and solutions refined through human curation based on solutions generated by LRMs such as DeepSeek-R1.
Most importantly, it includes only problems with verifiable answers, limited to integers within three digits.

\subsection{Supervised Fine-tuning}
\label{app:sft}
We employ two base models: Qwen-2.5-7B-Instruct and Llama-3.1-8B-Instruct.
The models are trained on the long CoT collection using 4 A100 GPUs.
We adopt LLaMA-Factory~\citep{zheng2024llamafactory}, a unified framework that integrates a suite of cutting-edge efficient training methods, to efficiently train the models.\footnote{\url{https://github.com/hiyouga/LLaMA-Factory}}
Detailed hyperparameters used for the training are provided in Table~\ref{tab:hyperparameter_sft}.

\subsection{RLVR}
\label{app:rlvr_hyperparameter}
Due to the limited GPU budgets, we employ Qwen-2.5-0.5B as a base model for training on long sequences with GRPO.
For our RL stage, we select a synthetic math dataset, NuminaMath~\citep{numina_math_datasets}, filtering problems based on Olympiads and AMC, resulting in a total of 10K problems.
We adopt OpenRLHF~\citep{hu2024openrlhf}, a framework designed to simplify and streamline RLHF training, and leverage RingAttnetion~\citep{liu2023ringattention} to enable training on long sequences.
Our RL stage is conducted on 16 A100 GPUs, and details about hyperparameters are in Table~\ref{tab:hyperparameter_rlvr}.


\subsection{Benchmark Details}
\label{app:benchmark}
AIME 2024 contains 30 problems administered on January 31–February 1, 2024. AIME assesses mathematical problem-solving across various domains including arithmetic, algebra, counting, geometry, number theory, and probability. 
MATH~\citep{hendrycks2021measuringmathematicalproblemsolving} comprises competition mathematics problems spanning different difficulty levels. Following previous work by OpenAI~\citep{lightman2023letsverifystepstep}, we use the same subset of 500 problems for evaluation.
Along with the mathematical benchmarks, we test our model on the general reasoning benchmarks, GPQA Diamond~\citep{gpqa}, a dataset consists of 198 doctorate-level questions across Biology, Chemistry, and Physics, and MMLU-Pro~\citep{wang2024mmlu} an enhanced version of MMLU~\citep{hendrycks2021measuringmassivemultitasklanguage} with a stronger focus on reasoning capabilities.

\subsection{Inference}
\label{app:inference}

all experiments are conducted with a temperature of 0.6 and a maximum token length of 16K, except for BoN sampling.
For BoN, we use top-$p$ decoding with $p=0.95$ and $t=1.0$.
Each model generates $n$=1, 2, 4, 8, 16, and 32 responses on MATH-500 and AIME2024, and selects the one that contain correct answer.
Since we focus on reasoning tasks, where correct answer is clearly defined, the results of BoN are equal to Pass@$n$.
To efficiently test models across diverse benchmarks, we utilize Simple-Eval, an open-source library from OpenAI.\footnote{\url{https://github.com/openai/simple-evals}}

\begin{table}[t!]
\centering
\resizebox{0.85\linewidth}{!}{
    \begin{tabular}{ll}
    \toprule
    \textbf{Hyperparameters} & \textbf{Value} \\
    \midrule
    Base Model & Qwen-2.5-7B-Instruct / \\
     & Llama-3.1-8B-Instruct \\
    Torch dtype & BF16 \\
    Epoch & 3 \\
    Train Data & Long CoT Collection \\
    Learning Rate & 5e-6 \\
    Max Seq. Length & 8,192 \\
    Batch Size & 1 \\
    Gradient Accumulation & 8 \\
    \bottomrule
    \end{tabular}
}
\caption{Hyperparameters used in the supervised fine-tuning.}
\label{tab:hyperparameter_sft}
\end{table}
\begin{table}[t!]
\centering
\resizebox{0.85\linewidth}{!}{
    \begin{tabular}{ll}
    \toprule
    \textbf{Hyperparameters} & \textbf{Value} \\
    \midrule
    Base Model & Qwen-2.5-0.5B-LC \\
    Torch dtype & BF16 \\
    Epoch & 5 \\
    Train Data & NunimaMath-CoT \\
    Learning Rate & 5e-7 \\
    Max Seq. Length & 16,384 \\
    Batch Size & 64 \\
    Gradient Accumulation & 1 \\
    Samples per Prompt & 16 \\
    \bottomrule
    \end{tabular}
}
\caption{Hyperparameters used for GRPO}
\label{tab:hyperparameter_rlvr}
\end{table}


\begin{figure*}[!t]
    \centering
    \includegraphics[width=1.0\linewidth]{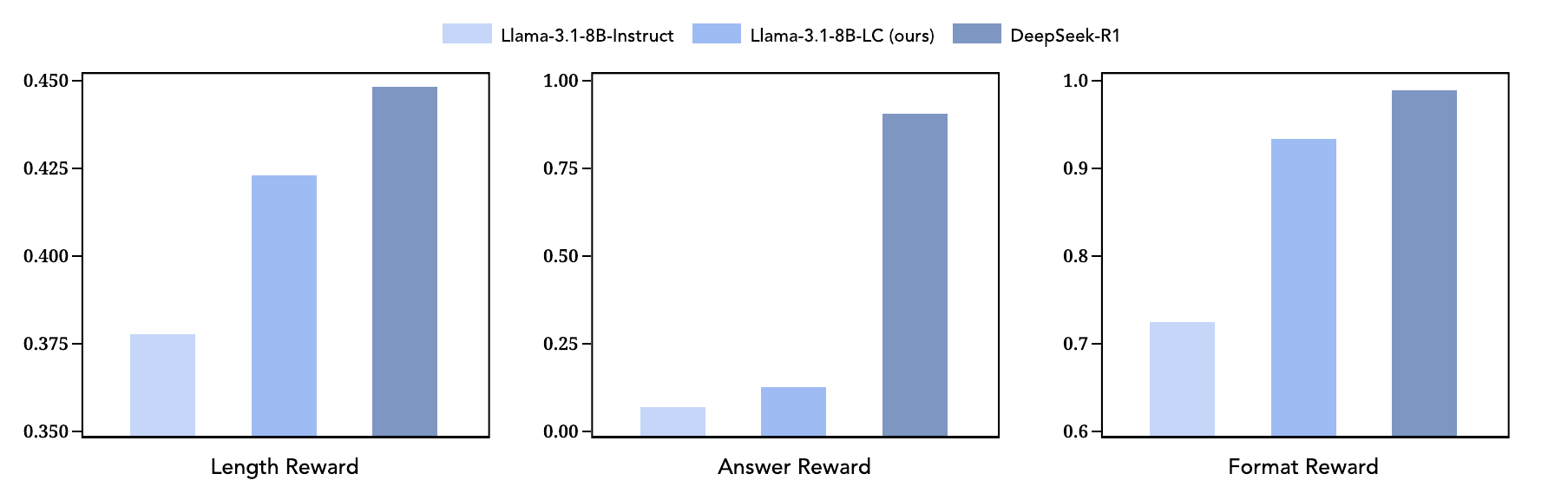}
    \caption{Length, answer, and format rewards across three models in LIMO dataset~\citep{ye2025limo}. 
}
    \label{fig:rewards}
\end{figure*}
\section{Impact on the Verifiable Rewards}
\label{ssec:impact_vr}
The success of RL is highly dependent on the SFT model~\citep{ouyang2022traininglanguagemodelsfollow}. 
We investigate the effect of initializing the SFT model with our dataset to the rewards for RL.
We utilize three reward functions aformentioned in Section~\ref{sec:implication_rl}.

Figure~\ref{fig:rewards} compares the averaged rewards of our model, Llama-3.1-8B-LC, with the baselines, Llama-3.1-8B-Instruct and R1. 
Among the three models, our model shows the highest length reward, suggesting the effectiveness of our dataset construction process in efficiently allocating thought tokens. Furthermore, our model's higher answer reward compared to Llama-3.1-8B-Instruct indicates its potential as an effective starting point for RL.


\section{Details on Analyses}
\label{app:analyses_details}

\subsection{Comparison on Thought Budget}
To compare thought budgets across different models, we employ model-specific token counting methods. For OpenAI's LRMs, we calculate the thought tokens by subtracting the response sequence tokens from the total completion tokens provided in the API response. For R1, which provides complete responses, we extract the content between \texttt{<think>} and \texttt{</think>} tags and count tokens using the GPT tokenizer. Similarly, for our model's responses, we measure the token count of sequences within the \texttt{<thought>} and \texttt{</thought>} tags.
\label{app:thought_budget}

\subsection{Details of the CoT Quality Analyses}
We use o3-mini as a judge and ask the model to identify which reasoning path is better based on the given criteria. The model chooses among the available options - A, B, or tie - where the two models' responses are randomly assigned to A and B for unbiased comparison.
\label{app:qualitative_analyses}


\section{Examples of the Long CoT Collection}
\label{app:example_long_cot_collection}

\begin{figure*}
    \centering
    \includegraphics[width=1.0\linewidth]{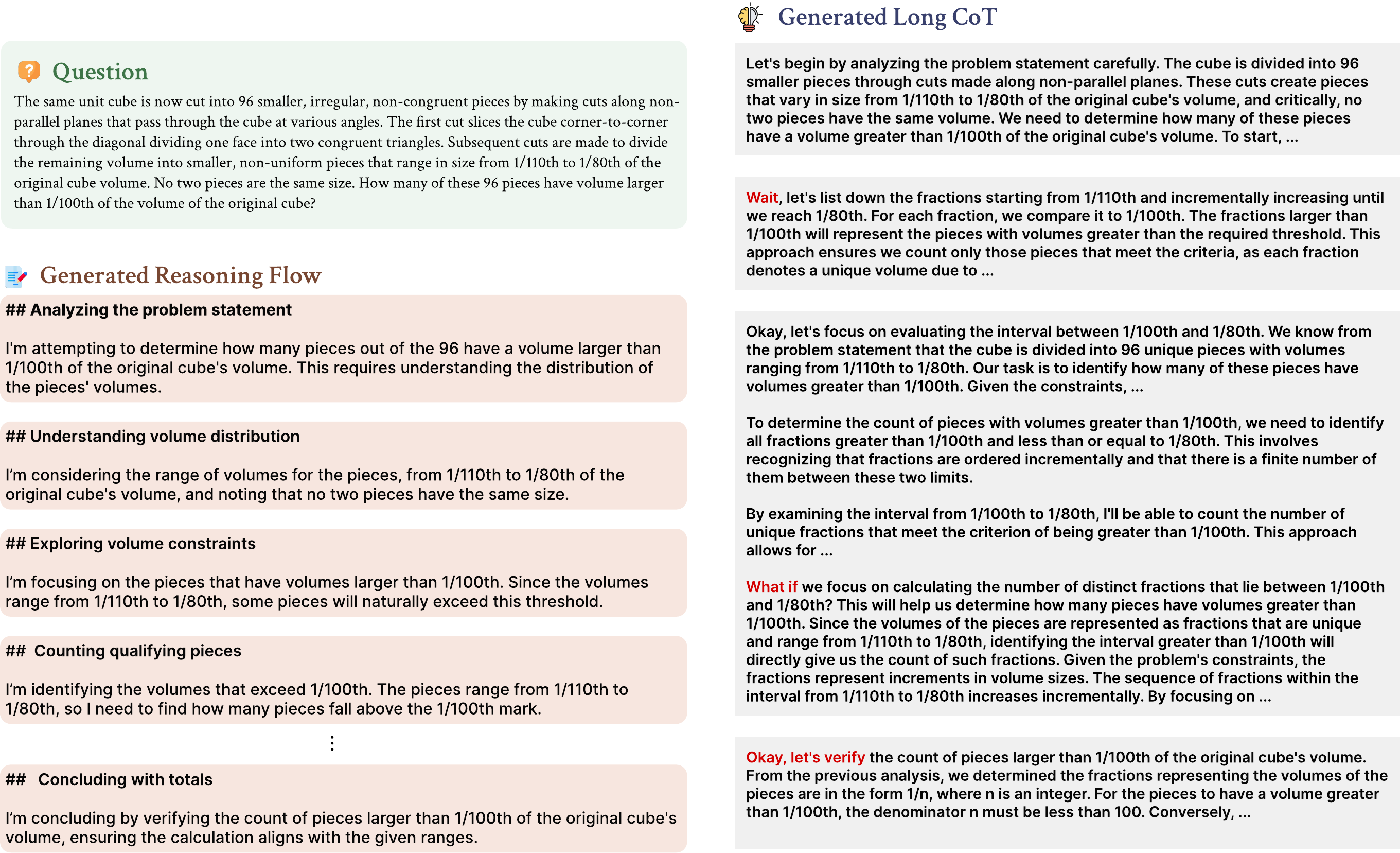}
    \caption{An example instance from the Long CoT Collection.}
    \label{fig:longcot_example2}
\end{figure*}

\begin{figure*}
    \centering
    \includegraphics[width=0.7\linewidth]{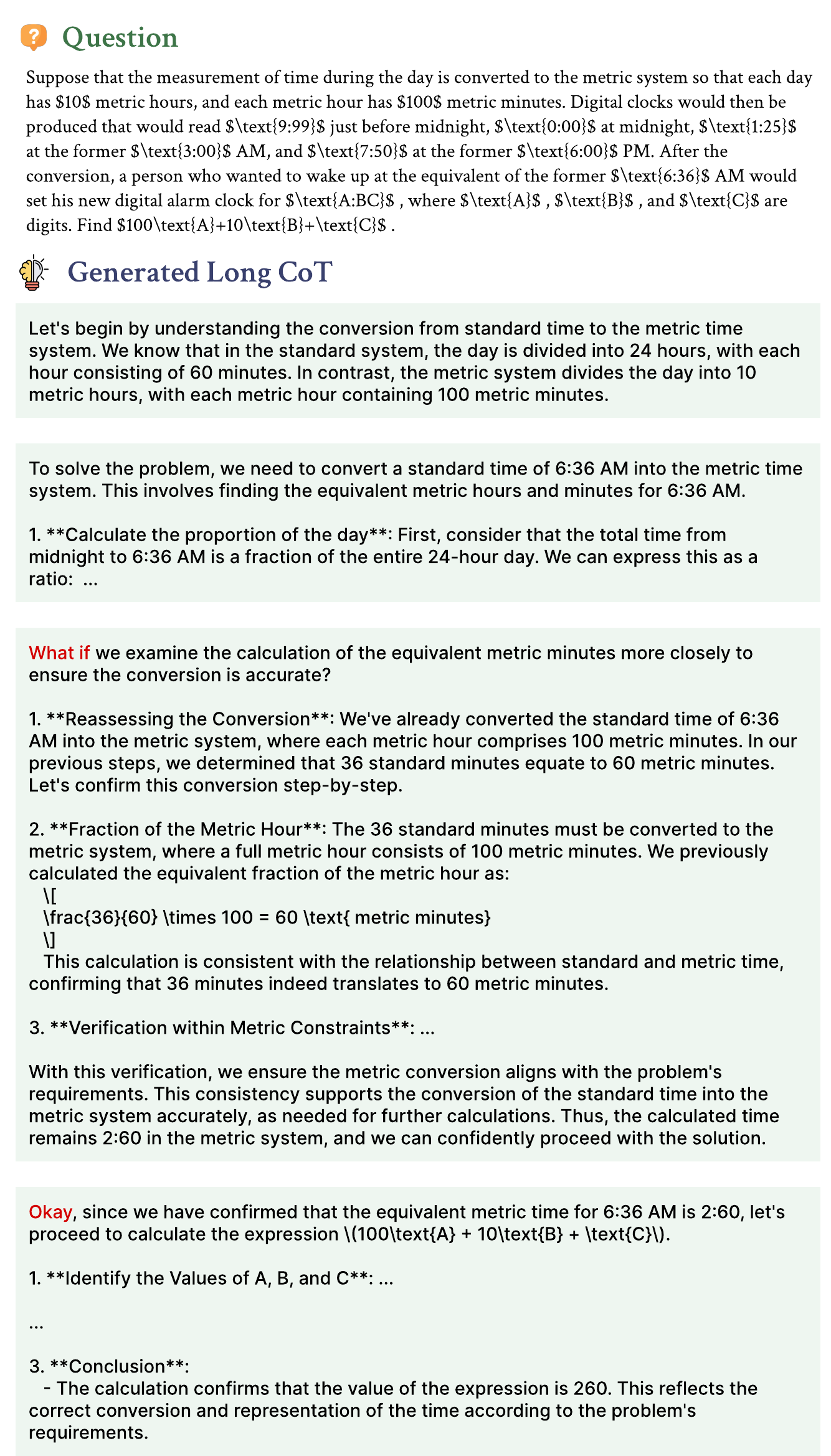}
    \caption{An example response from Llama-3.1-8B-LC (Ours), which trained on the Long CoT Collection.}
    \label{fig:example_ours}
\end{figure*}

We provide several examples from the Long CoT Collection:
\begin{itemize}
    \item An example of our Long CoT Collection Figure~\ref{fig:longcot_example2}
    \item An example response of Llama-3.1-8B-Instruct (Ours), which trained on the Long CoT Collection: Figure~\ref{fig:example_ours}
\end{itemize}


\section{Prompts}
\label{app:prompts}

\begin{figure*}[p]
    \centering
    \includegraphics[width=0.85\linewidth]{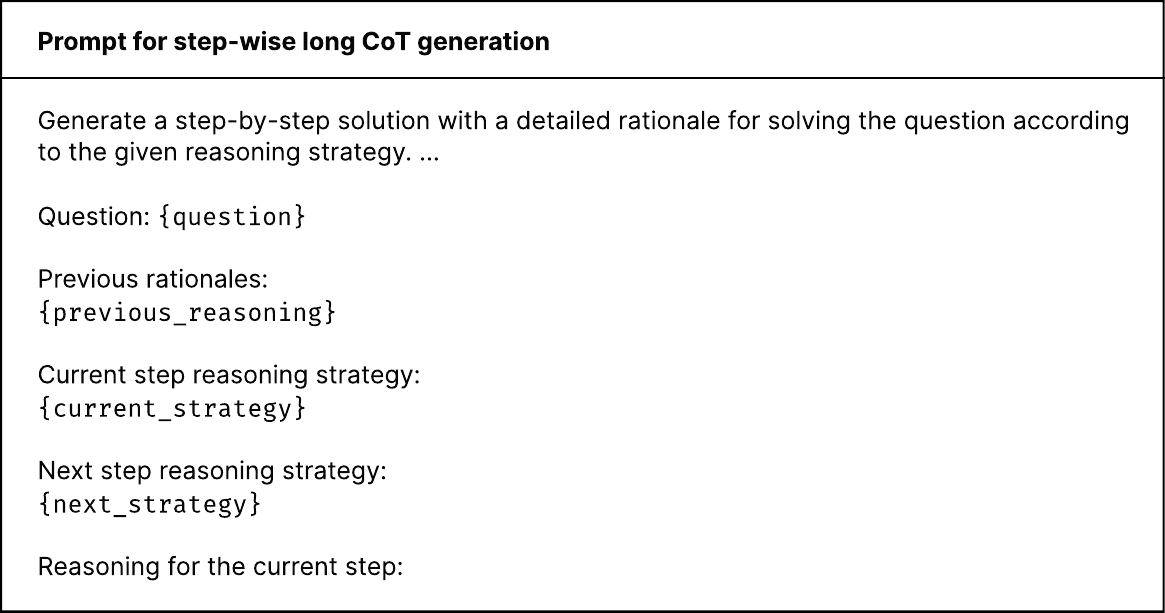}
    \caption{Prompt used for stepwise long CoT generation.}
    \label{fig:prompt_cot_generation}
\end{figure*}

\begin{figure*}[p]
    \centering
    \includegraphics[width=1\linewidth]{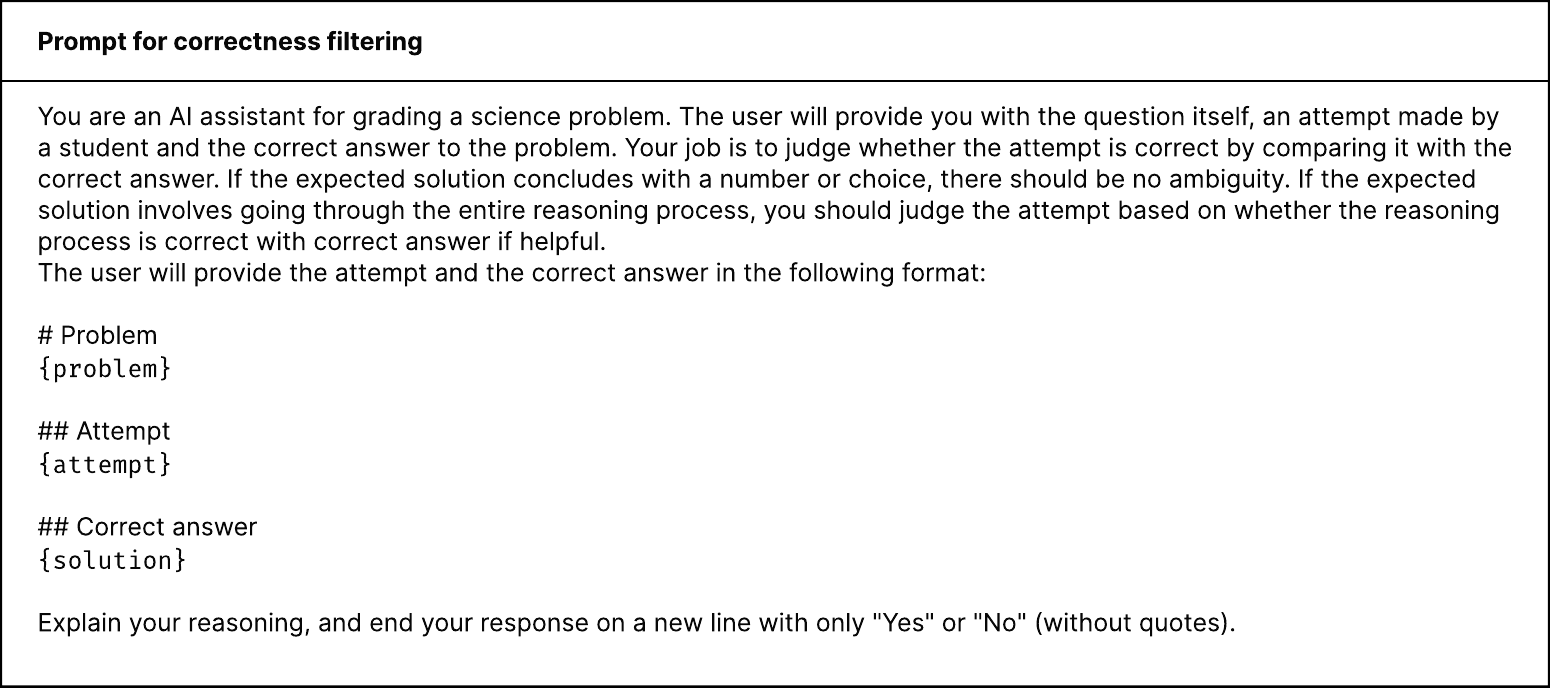}
    \caption{Prompt used for filtering the incorrect rationales.}
    \label{fig:prompt_filtering}
\end{figure*}

\begin{figure*}[htbp]
    \centering
    \includegraphics[width=0.7\linewidth]{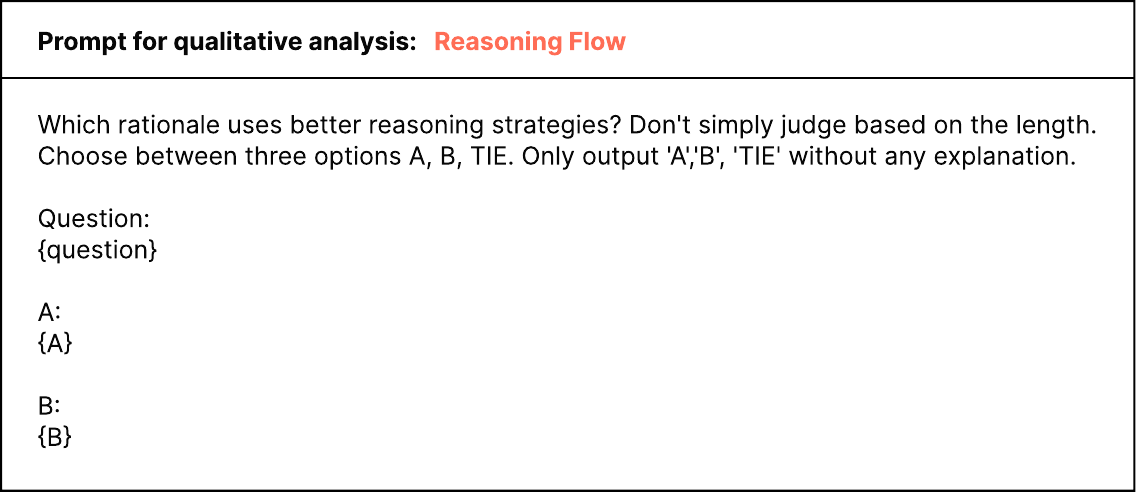}
    \caption{Prompt used for qualitative analysis on Reasoning Flow. We assign the position of A/B randomly.}
    \label{fig:prompt_reasoning_flow}
\end{figure*}

\begin{figure*}[htbp]
    \centering
    \includegraphics[width=0.7\linewidth]{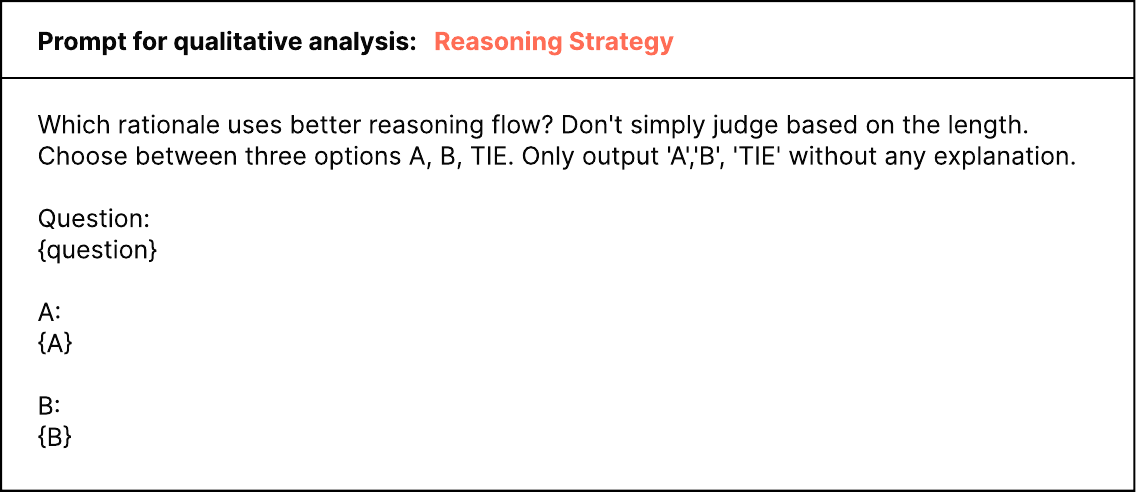}
    \caption{Prompt used for qualitative analysis on Reasoning Strategy. We assign the position of A/B randomly.}
    \label{fig:prompt_reasoning_strg}
\end{figure*}

\begin{figure*}[htbp]
    \centering
    \includegraphics[width=0.7\linewidth]{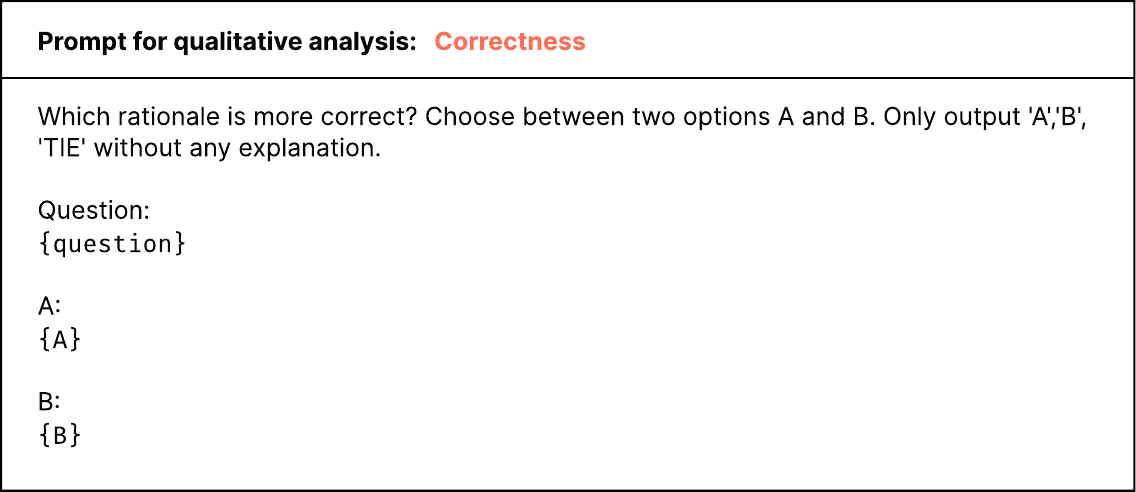}
    \caption{Prompt used for qualitative analysis on Correctness. We assign the position of A/B randomly.}
    \label{fig:prompt_correctness}
\end{figure*}

These are the prompts we utilized in our study:
\begin{itemize}
    \item Prompt for the step-wise long CoT generation: Figure~\ref{fig:prompt_cot_generation}
    \item Prompt for the correctness filtering: Figure~\ref{fig:prompt_filtering}
    \item Prompt for the CoT quality analyses in Section~\ref{ssec:high_quality}: Figure~\ref{fig:prompt_reasoning_flow},~\ref{fig:prompt_reasoning_strg}, and~\ref{fig:prompt_correctness}.
\end{itemize}

\section{Usage of AI Assistant}
We used ChatGPT for simple grammar correction and paraphrasing our draft.

\section{Artifact Licenses}
\begin{itemize}
    \item \textbf{magpie-reasoning-V1 dataset}: META LLAMA 3 COMMUNITY LICENSE AGREEMENT
    \item \textbf{AIME2024}: MIT license
    \item \textbf{MATH-500}: MIT license
    \item \textbf{LIMO dataset}: MIT license
    \item \textbf{GPQA diamond dataset}: cc-by-4.0
\end{itemize}


\begin{figure}[t]
    \centering
    \includegraphics[width=1.0\linewidth]{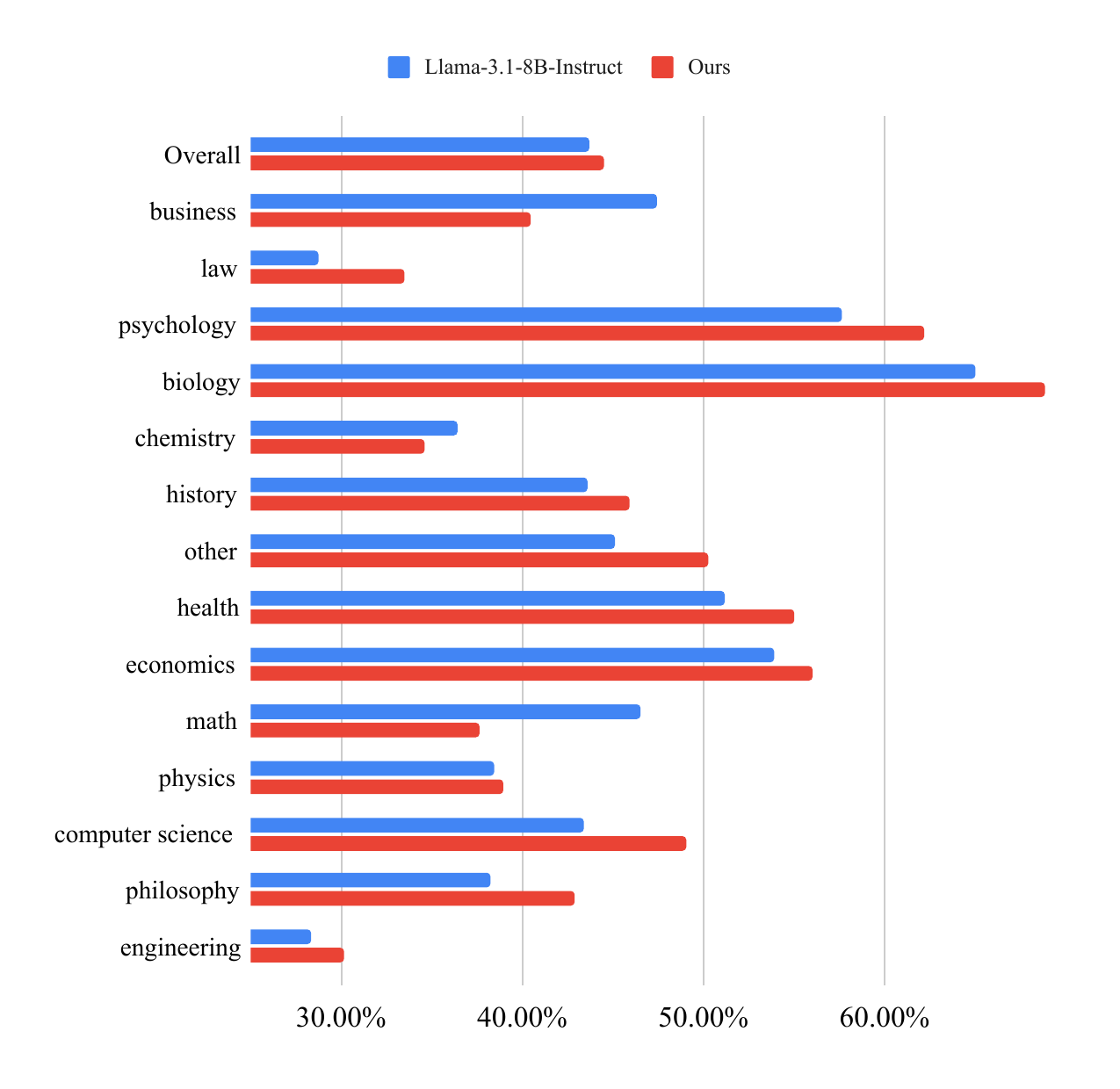}
    \caption{Results of Llama-3.1-8B-LC on MMLU-Pro broken down by domain.}
    \label{fig:mmlu_pro_llama}
\end{figure}
\begin{figure}[t]
    \centering
    \includegraphics[width=1.0\linewidth]{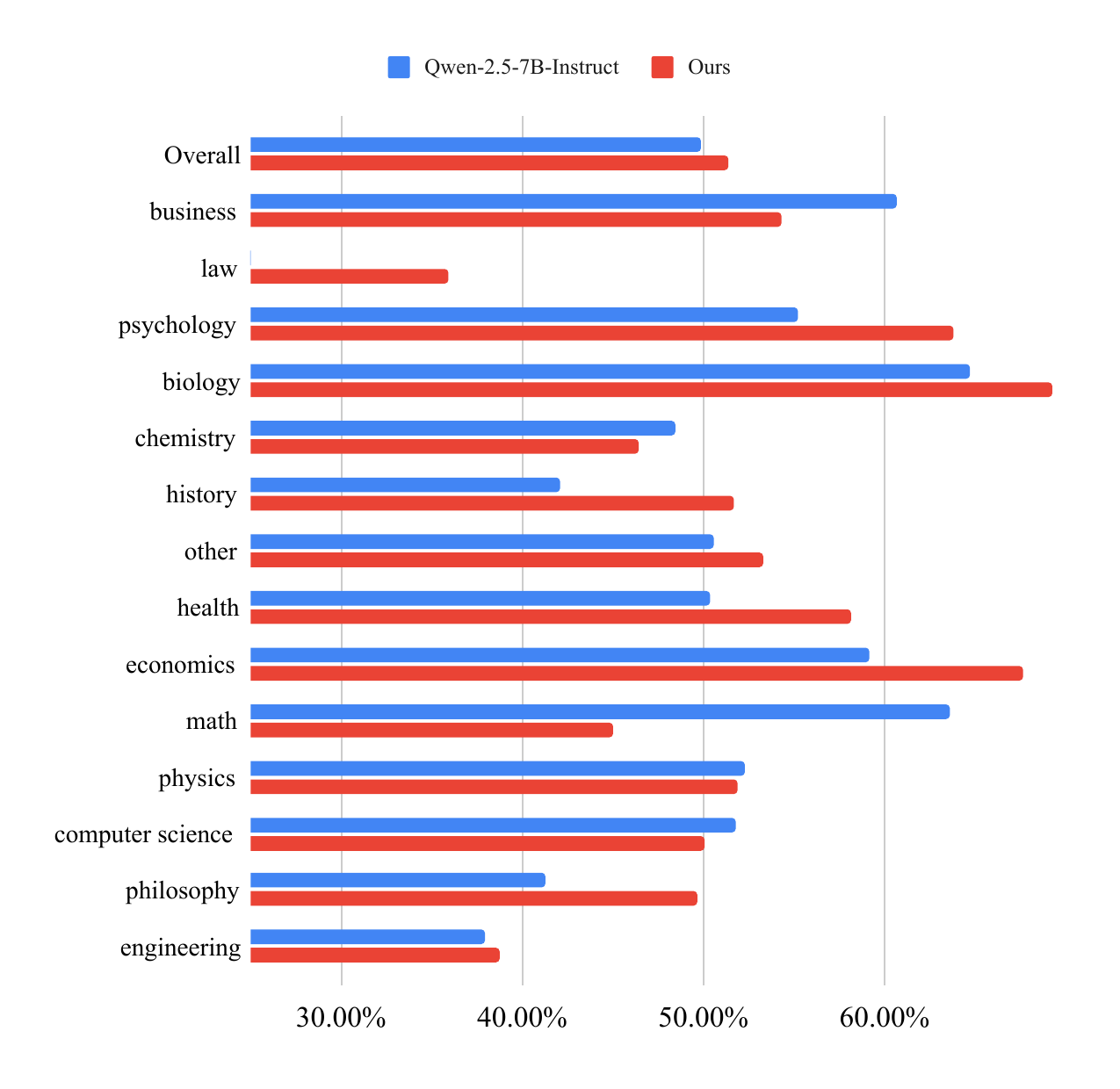}
    \caption{Results of Qwen-2.5-7B-LC on MMLU-Pro broken down by domain.}
    \label{fig:mmlu_pro_qwen}
\end{figure}

\begin{figure}[t]
    \vspace{-5px}
    \centering
    \includegraphics[width=0.9\linewidth]{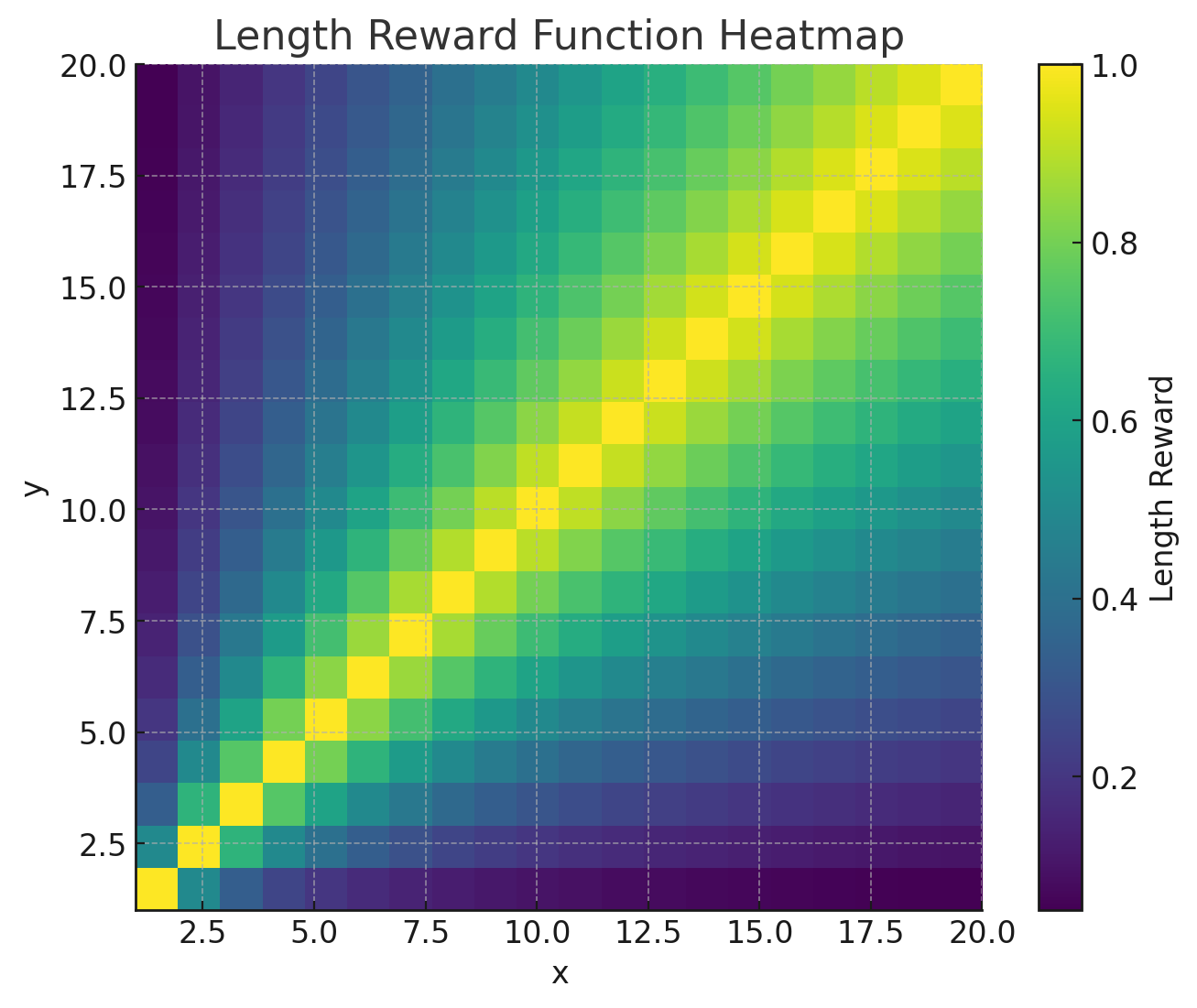}
    \caption{Heatmap of the thought budget function, defined as $1 - \left| \frac{\min(x, y)}{\max(x, y)} - 1 \right|$, where $x$ and $y$ are positive integers. Brighter regions indicate higher rewards, which occur when $x$ and $y$ are closer in value.}
    \label{fig:length_reward}
    \vspace{-5px}
\end{figure}

\clearpage

\clearpage

\end{document}